\documentclass{article}
\usepackage{ijcai23}
\usepackage{multirow}
\usepackage{diagbox}
\usepackage{times}
\usepackage{soul}
\usepackage{url}
\usepackage{amsmath}
\usepackage{amssymb}
\usepackage{mathptmx}
\usepackage[hidelinks]{hyperref}
\usepackage[utf8]{inputenc}
\usepackage[small]{caption}
\usepackage{graphicx}
\usepackage{amsmath}
\usepackage{amsthm}
\usepackage{booktabs}
\usepackage{algorithm}
\usepackage{algorithmic}
\usepackage[switch]{lineno}
\usepackage{float}

\title{TIM: An Efficient Temporal Interaction Module for Spiking Transformer}

\author{
Sicheng Shen$^1$$^,$$^2$$^,$$^4$\thanks{equal contribution}
\and
 Dongcheng Zhao$^1$$^,$$^2$\footnotemark[1]\and
Guobin Shen$^1$$^,$$^2$$^,$$^4$\and
Yi Zeng$^1$$^,$$^2$$^,$$^3$$^,$$^4$$^,$$^5$\thanks{corresponding author}
\affiliations
$^1$ Brain-inspired Cognitive Intelligence Lab, Institute of Automation, Chinese Academy of Sciences\\
$^2$ Center for Long-term Artificial Intelligence \\
$^3$ Key Laboratory of Brain Cognition and Brain-inspired Intelligence Technology, CAS \\
$^4$ School of Future Technology, University of Chinese Academy of Sciences\\
$^5$ School of Artificial Intelligence, University of Chinese Academy of Sciences\\
\emails
\{shensicheng2024, zhaodongcheng2016, shenguobin2021, yi.zeng\}@ia.ac.cn
}

\begin{document}

\maketitle

\begin{abstract}
    Spiking Neural Networks (SNNs), as the third generation of neural networks, have gained prominence for their biological plausibility and computational efficiency, especially in processing diverse datasets. The integration of attention mechanisms, inspired by advancements in neural network architectures, has led to the development of Spiking Transformers. These have shown promise in enhancing SNNs' capabilities, particularly in the realms of both static and neuromorphic datasets. Despite their progress, a discernible gap exists in these systems, specifically in the Spiking Self Attention (SSA) mechanism's effectiveness in leveraging the temporal processing potential of SNNs. To address this, we introduce the Temporal Interaction Module (TIM), a novel, convolution-based enhancement designed to augment the temporal data processing abilities within SNN architectures. TIM's integration into existing SNN frameworks is seamless and efficient, requiring minimal additional parameters while significantly boosting their temporal information handling capabilities. Through rigorous experimentation, TIM has demonstrated its effectiveness in exploiting temporal information, leading to state-of-the-art performance across various neuromorphic datasets. The code is available at \href{https://github.com/BrainCog-X/Brain-Cog/tree/main/examples/TIM}{https://github.com/BrainCog-X/Brain-Cog/tree/main/examples/TIM}.
\end{abstract}

\section{Introduction}


Spiking neural networks (SNNs), representing a novel paradigm in the evolution of artificial neural networks (ANNs), derive their foundational principles from the intricacies of biological neural systems, with a particular focus on dynamic and temporal processing capabilities~\cite{maass1997networks,zeng2023braincog}. The event-driven mechanism inherent in SNNs markedly amplifies their energy efficiency, while simultaneously fostering improved interoperability with a broad range of neuromorphic hardware~\cite{roy2019towards,wei2024event}. These networks have emerged as a pivotal tool within the domain of cognitive and intelligent systems research.

In the development of SNNs,  researchers initially borrowed structures from biological neural systems but found challenges in scaling these structures for large networks and diverse tasks~\cite{cheng2020lisnn,zhao2020glsnn,dong2023unsupervised}. To overcome these limitations, well-established designs from the deep learning domain, like ResNet and VGGNet, were introduced to improve flexibility and applicability~\cite{sengupta2019going,fang2021deep,shen2022backpropagation}. In addition, the Transformer architecture, known for its excellent parallel processing and ability to handle long-distance dependencies, has become prominent in various domains~\cite{vaswani2017attention,dosovitskiy2021vit}. Its self-attention mechanism offers unmatched flexibility and efficiency in handling sequential data. Currently, there's growing interest in combining SNNs with Transformers, using the biological realism and energy efficiency of SNNs alongside the powerful data processing of Transformers, opening up new prospects for simulating complex cognitive functions~\cite{zhou2022spikformer,li2022spikeformer}.

The rapid advancement of event-based cameras has heralded a new era in the realms of image sensing and computer vision~\cite{gallego2020event,zheng2023deep,zhou2023computational,rebecq2019high,gehrig2019end}. Diverging from traditional imaging, event streams store information in the form of events rather than pixels. This approach not only preserves the temporal aspects of images to a certain extent but also reduces the resources required for image processing. Consequently, event streams are regarded as a more biologically congruent, efficient, and promising method for image storage and processing. The integration of SNNs with Transformers offers a highly efficient and adaptable strategy for handling event-driven data. Leveraging the biological principles underlying SNNs, this combination exhibits exceptional performance in dynamic and sequential data processing. Transformers enhance this synergy with their capability for efficient parallel processing and precise handling of long-range dependencies, thereby optimizing the analysis of complex data. This fusion not only emulates the processing mechanisms of biological neural systems but also significantly elevates the efficiency and accuracy of data processing. Particularly in the processing of event-based camera data, this amalgamation, utilizing spike-based event stream storage, effectively retains temporal information while concurrently reducing resource consumption.

Spikformer~\cite{zhou2022spikformer} represents the first successful integration of the Transformer architecture into the SNN domain. This model innovatively designs Spiking Self Attention to implement Transformer attention. In recent studies of Spiking Transformers, many improvements have been made. Currently, there are two primary approaches to improving Spiking Transformers. The first approach involves enhancing network performance by modifying the attention mechanism (eg. Spikeformer~\cite{li2022spikeformer} and DISTA~\cite{xu2023dista}). The second focuses on leveraging the efficiency of SNNs to reduce the computational energy consumption of Transformers (e.g. SPikingformer~\cite{zhou2023spikingformer} and Spike-diriven Transformer~\cite{yao2023spikedriven}). However, these advancements encounter limitations when applied to neuromorphic data. The unique dynamics and temporal complexity of neuromorphic datasets, reflecting biological neural systems, challenge these models. While they enhance network performance in some aspects, they fall short in fully capturing the nuances of neuromorphic data processing, highlighting a need for further specialized adaptations in Spiking Transformers to effectively manage these specific data characteristics.

\begin{figure*}[t]
  \centering
  \graphicspath{{images/}}
  \includegraphics[width=1\textwidth]{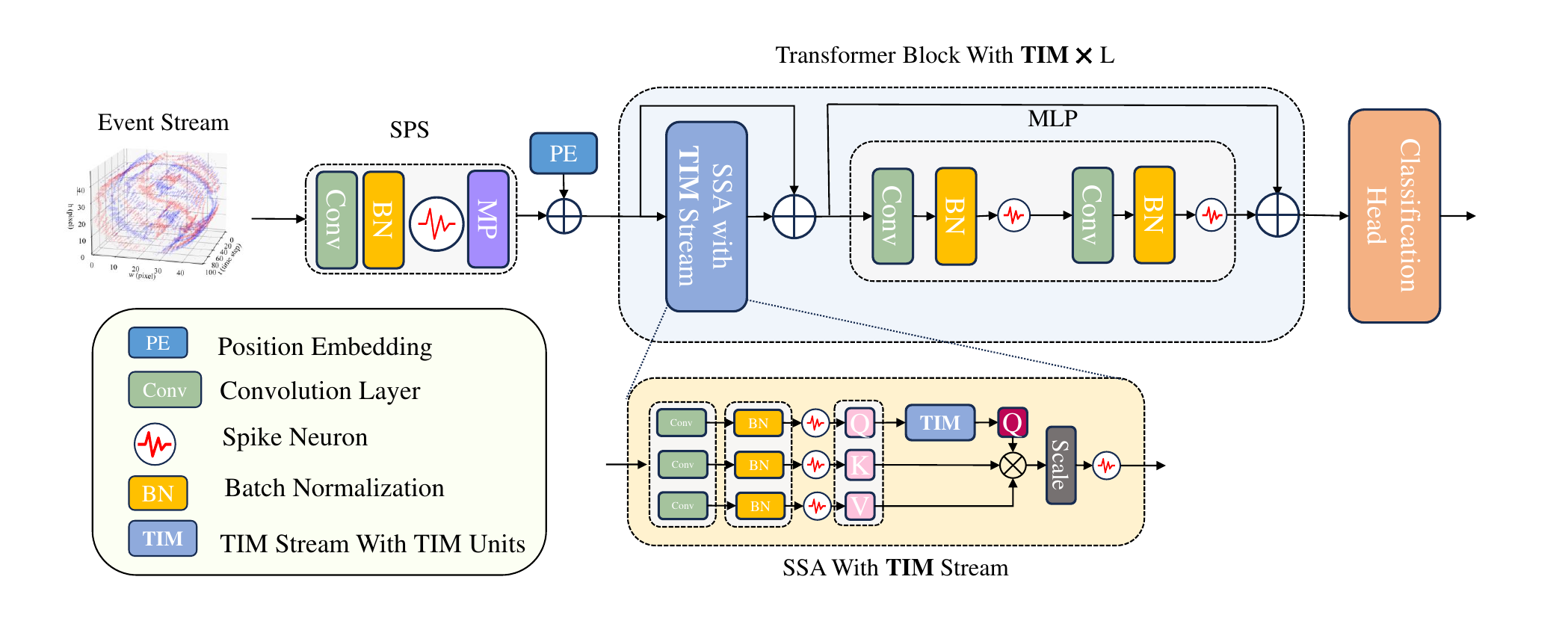}
  \caption{Comprehensive diagram of Spikformer integrated with Temporal Interaction Module (TIM): Demonstrating TIM's Plug-and-Play role within the Spike Self Attention (SSA) Block of the original Spikformer structure.}
  \label{fig:pipline}
\end{figure*}

In this study, we have developed a Temporal Interaction Module that can be seamlessly integrated into the attention matrix computation of a Spiking Transformer. This module enables the adaptive amalgamation of historical and current information, thereby effectively capturing the intrinsic dynamics of neuromorphic data. Our experiments conducted across various neuromorphic datasets demonstrate that our approach achieves the state-of-the-art performance of Spiking Transformer. In summary, our contributions are as follows:
\begin{itemize}
  \item Through our analysis, we identified that a primary limitation in current Spiking Transformers is their insufficient handling of temporal information, chiefly because the attention matrix is reliant solely on information from the current moment.
  \item To address this, we have designed a Temporal Interaction Module to adaptively utilizes information from different time steps and functions as a lightweight addition. It can be integrated with existing attention computation modules without significantly increasing computational load.
  \item We conducted experiments on neuromorphic datasets, including CIFAR10-DVS, NCALTECH101, NCARS, UCF101-DVS, HMDB-DVS and SHD. We demonstrated the effectiveness and generalization ability of our method. Our results show that our approach sets a new benchmark in the Spiking Transformer domain, achieving better performance across these datasets.
\end{itemize}

\section{Related Work}
In this section, we will review and analyze recent research aimed at enhancing the temporal information processing capabilities of SNNs and developments in the Spiking Transformer field.
\subsection{Advances in SNN Temporal Processing}
In terms of enhancing the temporal information processing ability of SNNs, significant progress has been made by researchers. For instance, ~\cite{kim2023exploring} explored the dynamic characteristics of temporal information in SNNs by estimating the Fisher information of weights. PLIF~\cite{fang2021incorporating} enhanced the integration of information at different time steps in SNNs by introducing a learnable membrane potential constant. IIRSNN~\cite{fang2020exploiting} improved the spatiotemporal information processing capabilities of SNNs through synaptic modeling.~\cite{zhang2021rectified} discovered that learning based on spike timing in an event-driven manner can yield significant improvements in classification accuracy.TA-SNN~\cite{yao2021temporal} introduced a temporal-based attention mechanism, enabling the adaptive allocation of importance to different time steps. TCJA~\cite{zhu2024tcja} incorporates an attention mechanism designed to assess the significance of pulses in both temporal and spatial dimensions.~\cite{shen2024exploiting} significantly bolstered the temporal information processing capability of SNNs by incorporating dendritic nonlinear computations. ETC~\cite{zhao2023improving} introduced Temporal Enhancement Consistency constraints to enable SNNs to learn from outputs at different time steps, thereby enhancing performance and reducing latency. BioEvo~\cite{shen2023brain} employed a biologically inspired neural circuit search approach, adaptively coordinating different circuits to improve the performance of SNNs in perceptual tasks and reinforcement learning.
\subsection{Spiking Transformer}
In the Transformer domain, Spikformer~\cite{zhou2022spikformer} innovatively implemented an attention mechanism called Spiking Self Attention. This mechanism replaced the traditional activation function of ANNs with spiking neurons and omitted the softmax function typically required before calculating attention~\cite{vaswani2017attention}, opting instead for direct matrix multiplication. DISTA~\cite{xu2023dista} enhanced the capture of spatiotemporal data by designing new neuronal connections. Spikeformer~\cite{li2022spikeformer} introduced separate Temporal and Spatial Multi-head Self Attention in each Transformer block to strengthen data processing capabilities. Spikingformer~\cite{zhou2023spikingformer} achieved greater efficiency by rearranging the positions of neurons and convolutional layers to eliminate float-integer multiplications. In contrast, the Spike-driven Transformer~\cite{yao2023spikedriven} reduced time complexity by altering the order of attention computation and substituting multiplication with addition.

However, despite a series of advancements in the Spiking Transformer domain, these methods still exhibit performance shortcomings when processing information such as neuromorphic data with strong temporal characteristics. This limitation reveals the current technology's inadequacies in capturing and processing data with complex temporal dependencies, particularly in analyzing the deep temporal dynamics and subtle changes within neuromorphic data.

\section{Preliminaries}
Currently, the predominant approach in Spiking Transformer methodologies involves substituting the activation functions in conventional Transformer architectures with spiking neurons. In this section, we will provide an in-depth analysis of the limitations encountered during the implementation of this approach.
\subsection{Spiking Neuron}
Spiking neurons, serving as the fundamental computational units of SNNs, exhibit significant differences from conventional ANNs. Their distinctive feature lies in the temporal accumulation of membrane potential; a spike is emitted once this potential surpasses the threshold. Within the realm of SNNs, Leaky Integrate-and-Fire (LIF) neurons are extensively employed due to their optimal balance between computational efficiency and biological plausibility. To facilitate ease of simulation computations,  the discrete formulation of LIF neurons is used, and the details are shown in Eq.~\ref{lif1}.
\begin{equation}
      V[t] = V[t-1] + \frac{1}{\tau} \left( X[t] - (V[t-1]) \right)
      \label{lif1}
\end{equation}
where $\tau$  is the membrane time constant, $V[t]$  refers to the membrane potential of the neurons in the step t. $X[t]$ is the input in the step $t$. 

Following the emission of a spike, the membrane potential of a spiking neuron is reset to its resting potential $V_{\text{reset}}$. For computational convenience, this resting potential is set to zero. The details of this process are delineated in Eq.~\ref{lif2}:
\begin{equation}
      V[t] = V[t] \cdot (1 - \Theta(V[t]))
      \label{lif2}
\end{equation}
\begin{equation}
    \Theta(x) = 
    \begin{cases}
    0 & \text{if } x < V_{th} \\
    1 & \text{if } x \geq V_{th}
    \end{cases}
    \label{lif3}
\end{equation}
$\Theta(x)$ is the Heaviside function. $V_{th}$ is the firing threshold. 

To facilitate the application of the backpropagation algorithm for network training, we utilize surrogate gradients as an approximation for the gradients of the spike firing function. This method is implemented as follows:
\begin{equation}
\label{eq6}
\frac{\partial  \Theta}{\partial V} =\left\{
\begin{aligned}
& 0 , & |V-V_{th}| > \frac{1}{a} \\
& -a^2|V - V_{th}| + a  ,& |V-V_{th}| \leq \frac{1}{a}
\end{aligned}
\right.
\end{equation}

In our study, the variable $a$ serves as a hyperparameter, instrumental in dictating the configuration of the surrogate gradient's shape. We have strategically set the value of $a$ to 4. 
\subsection{Spiking Self Attention Analysis}
The primary strength of the Transformer model resides in its innovative self-attention mechanism. This mechanism facilitates a nuanced examination of the relative importance assigned to distinct positions within a sequence, consequently enhancing the model's capacity for sophisticated information processing. Building upon this foundational concept, the Spikformer model introduces an advancement with the Spike Self Attention (SSA) mechanism, a development that draws inspiration from Vaswani et al.'s seminal work~\cite{vaswani2017attention}. The intricacies of this mechanism are delineated in Eq.~\ref{eq:SSA}.
\begin{equation}
\left\{
    \begin{aligned}
        Q[t],K[t],V[t] &= LIFNode(X[t]) \\
        A[t] &= Q[t]K[t]^TV[t]
    \end{aligned}
\right.
\label{eq:SSA}
\end{equation}
Upon the amalgamation of SSA  and LIF , it is deducible that the conduit of information transmission within the spiking Transformer architecture is characterized by Eq.~\ref{lif4}:
\begin{align}
    V[t] &= V[t-1] + \frac{1}{\tau} \left( A[t] - (V[t-1]) \right)\notag \\
           & = (1-\frac{1}{\tau})V[t-1] + \frac{1}{\tau}A[t]  
    \label{lif4}
\end{align}

\begin{figure*}[t]
  \centering
  \graphicspath{{images/}}
  \includegraphics[width=1\textwidth]{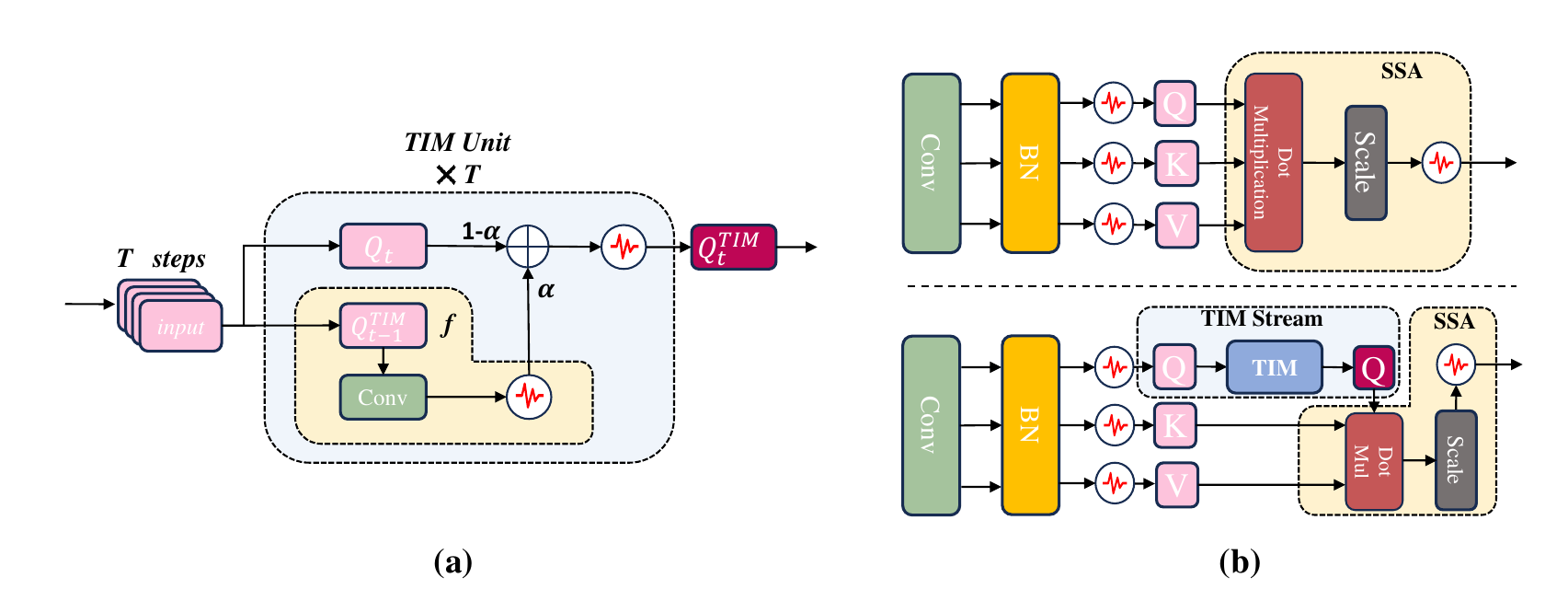}
  \caption{(a): Overview of the Temporal Interaction Module: Adaptive Integration of Historical and Current Temporal Data. (b): The upper part of the figure describes the process of SSA, while the lower part illustrates the process after integrating TIM Stream into SSA. }
  \label{fig:unit_stream}
\end{figure*}
It is apparent that the membrane potential at a given time, $V[t]$, is principally determined by its previous state $V[t-1]$ and the current input $A[t]$. The operation $Q[t]K[t]^TV[t]$ is chiefly responsible for facilitating the interaction of spatial information. However, the retention and extraction of temporal information rely solely on the dynamic changes in the neuronal membrane potential. In the existing spike self-attention mechanisms, temporal information has not been adequately considered, a deficiency that significantly limits the spiking Transformer's capability in processing time-series information.

\section{Method}
To enhance the temporal information processing capability of the Spiking Transformer, we have meticulously developed a plug-in \underline{T}emporal \underline{I}nteraction \underline{M}odule (TIM). In this section, we will detail the architecture of TIM and explore its integration and application within the SSA computation.

\subsection{Spikformer Backbone}
In our experiments employing the Temporal Interaction Module, we have chosen Spikformer~\cite{zhou2022spikformer} as the primary framework, with its network architecture comprehensively illustrated in Fig.~\ref{fig:pipline}. Given TIM's emphasis on temporal enhancement within neuromorphic datasets, our inputs are formatted as Event Streams. We have preserved the key components of Spikformer for consistency across our experimental models. The Event Stream undergoes a transformation into requisite dimensions through the Spiking Patch Splitting (SPS) process, utilizing convolutional techniques. Within the Spiking Self Attention (SSA) module, the TIM Stream is employed for query operations to facilitate temporal enhancement. The Multi-Layer Perceptron (MLP) is structured with a bi-layered Convolution-Batch Normalization-Leaky Integrate-and-Fire (Conv-BN-LIF) framework. Finally, the Classification Head is constituted by a linear layer, aligning with the model's output objectives.

\subsection{TIM Unit}

The effective preservation, processing, and extraction of temporal features are crucial in handling neuromorphic data. As demonstrated in Eq.~\ref{eq:SSA}, traditional Spiking Transformers construct an attention matrix at each time step. However, this attention mechanism solely correlates with the current input, leading to a substantial underutilization of information from different time steps.


Here, $f$ signifies an operation for extracting historical information, implemented as a one-dimensional convolution in this study. Consequently, $Q^{TIM}[t]$ comprises two components: the first represents the contribution of historical information to the current attention, while the second signifies the direct impact of the current input. The hyperparameter $\alpha$ facilitates an adaptive balance between the significance of historical and current information. The specific illustration of TIM Unit can be found in Fig~\ref{fig:unit_stream}(a).


\subsection{TIM Stream}
As depicted in Fig.~\ref{fig:unit_stream}(b), this module is integrated into the computational graph of the attention matrix.
\begin{align}
    &Attention[t] = A^{TIM}[t]K[t]^TV[t] \\
            &= \alpha f(Q^{TIM}[t-1])K[t]^TV[t] + (1- \alpha)Q[t]K[t]^TV[t] \notag  
    \label{TIM2}
\end{align}
Compared to the traditional SSA, our introduction of the $f$ operation, which employs a one-dimensional convolution, does not significantly increase the number of parameters. By incorporating the TIM-built attention matrix, the model is capable of not only processing information from the current moment but also of utilizing output information from past moments, thereby effectively capturing the intrinsic dynamics of time series. This enhancement significantly bolsters the model's temporal memory capability and allows it to utilize historical information at each time step. This feature enables the model to dynamically adjust its behavior based on the characteristics of different tasks and data, thus improving computational efficiency and the generalization capacity of the model.
\section{Experiments}
To validate the performance of our algorithm, we conducted comprehensive tests on multiple neuromorphic datasets, including DVS-CIFAR10, N-CALTECH101, NCARS, UCF101-DVS, and HMDB51-DVS. Furthermore, to highlight the exceptional performance of our algorithm, we compared it with the current leading SNN models. All experiments were completed on the BrainCog~\cite{zeng2023braincog} platform. In the experiments, we set the batchsize to 16 and used the AdamW optimizer. The total number of training epochs was set to 500. The initial learning rate was set to 0.005, adjusted with a cosine decay strategy. The time constant ($\tau$ value) of the LIF Node was set to 2, and its firing threshold was set to 1. The simulation step length of the SNN was set to 10. The default $\alpha$ of the TIM Stream was set to 0.5.

\begin{table*}[!t]
    \centering
    \resizebox{0.95\textwidth}{!}{
    \begin{tabular}{llllll}
        \hline
        \textbf{Method}  & \textbf{Architecture} &  \textbf{Steps} &  \textbf{CIFAR10-DVS} &\textbf{N-CALTECH101} &\textbf{NCARS} \\
        \hline
        SALT~\cite{kim2021optimizing}     & VGG-11  & 20  &67.1 &55.0 &- \\
         Rollout~\cite{kugele2020efficient}  & VGG-16     & 48    &66.5 &- &94.1 \\
        SEW-ResNet~\cite{fang2021deep}     & Wide-7B-Net    & 20  &74.4&- &- \\
        ResNet-18~\cite{shen2022eventmix} &ResNet-18 &10 &79.2& 75.3 & 95.9 \\
        EventMix~\cite{shen2022eventmix} & ResNet-18 &10 &81.5 &79.5 &96.3 \\
        DT-SNN~\cite{li2023input}  & ResNet-19 & 10   & 74.8 &- &-\\
        NDA~\cite{li2022neuromorphic} & ResNet-19 & 10   & 78.0 &78.6 &87.2\\
        Event Transformer~\cite{li2022event}  &Transformer &-\textsuperscript{*} &71.2 &78.9 &95.4\\ 
        RM SNN~\cite{yao2023sparser}  &PLIF-SNN &10  &- &77.9 &-\\ 
        
        VMV-GCN~\cite{xie2022vmv} &VMV-GCN &-\textsuperscript{*} &69.0 &77.8 &93.2\\
        \hline
        Spikformer~\cite{zhou2022spikformer} & Spikformer & 10 / 16 & 78.9 / 80.9  &- &- \\
        DISTA~\cite{xu2023dista} & Spikformer& 10  & 79.1 &- &- \\ 
        Spikingformer~\cite{zhou2023spikingformer} &Spikformer &10 / 16 & 79.9 / 81.3 &- &- \\
        Spike-driven 
        Transformer~\cite{yao2023spikedriven} &Spikformer & 16 & 80.0 &- &- \\
        2D-WT-Haar~\cite{wang2023attention} &Spikformer &16 &81.0 &- &-\\
        Spikeformer~\cite{li2022spikeformer} &Spikeformer &4 &81.4 &- &- \\
        \hline
       \textbf{TIM(Ours)} & Spikformer &  \textbf{10} & \textbf{81.6} & \textbf{79.0} & \textbf{96.5} \\
        \hline
    \end{tabular}
    }
    \caption{Comparison with other benchmark results on CIFAR10-DVS, N-CALTECH101 and NCARS. The $*$ indicates unknown time steps from the original studies.}
    \label{tab:dcs10ncalncars}
\end{table*}
\subsection{CIFAR10-DVS}

\textbf{CIFAR10-DVS}~\cite{li2017cifar10} is an event stream dataset comprising 10,000 images from the CIFAR-10 dataset. These images are converted into event streams using a bicubic interpolation method.  As outlined in Tab.~\ref{tab:dcs10ncalncars}, our algorithm exhibits a significant improvement over the prevalent Spiking Transformer framework. When benchmarked against Spikformer, our TIM demonstrates enhanced performance, exceeding it by approximately 2.7\% under identical configurations. Notably, even when Spikformer is optimized with a longer time step (Time Step=16), our algorithm maintains a performance lead of 0.7\%. While Spikeformer attains an accuracy of 81.4\% in 4 steps, it requires a substantial 9.28 million parameters. To our knowledge, this achievement marks the most advanced state-of-the-art (SOTA) result for Spiking Transformer applications on the CIFAR10-DVS dataset. In comparison with convolutional structures, the efficacy of TIM is on par with EventMix~\cite{shen2022eventmix}, which records a similar accuracy of 81.45\%. However, a striking advantage of our model is its efficiency: while EventMix operates with an extensive 11.69 million parameters, our TIM model achieves comparable results with a significantly reduced parameter count of only 2.59 million.
\subsection{N-CALTECH101}

The \textbf{N-CALTECH101} dataset, as introduced in~\cite{orchard2015converting}, is an innovative adaptation of the classic CALTECH101 image library, covering 101 diverse categories. This dataset is notably derived using advanced neuromorphic visual sensors. It uniquely offers a time-based visual event representation of the original static image categories, effectively capturing the dynamic and temporal nuances inherent in each category.  In our research, we have compiled and analyzed several benchmarks from the N-CALTECH101 dataset alongside the results of our Temporal Integration Model, as summarized in Tab.~\ref{tab:dcs10ncalncars}. These benchmarks primarily stem from ANN architectures. However, our SNN model based TIM exhibits a remarkable performance edge over these benchmarks. Despite being compared with the Event Transformer which is an ANN-based Transformer model, our SNN-based TIM has achieved comparable performance.

\subsection{NCARS}

The \textbf{NCARS} dataset, as introduced in~\cite{orchard2015converting}, represents a cutting-edge binary classification dataset. It consists of a comprehensive collection of dynamic event streams, meticulously capturing cars and non-car objects through the lens of event cameras. These event streams offer a rich, real-time depiction of visual phenomena, distinguishing themselves by their ability to encode temporal information about moving objects. Our TIM continues to set new benchmarks in the analysis of the NCARS dataset. It achieves a remarkable 96.5\% accuracy, a testament to its robustness and advanced feature extraction capabilities. This level of performance not only surpasses the Event Transformer by a notable margin of 1.1\% but also maintains a weak lead over EventMix, outperforming it by 0.2\%. These results underscore the efficacy of TIM in handling the complex dynamics and temporal variations inherent in the NCARS dataset.
\begin{table*}[t]
    \centering
    \resizebox{0.6\textwidth}{!}{
    \begin{tabular}{llll}
        \hline
        \textbf{Dataset}  & \textbf{Model} &\textbf{Acc@1(\%)}  \\ 
        \hline
                        
                        & Event Frames + I3D / 3D-ResNet~\cite{bi2020graph} & 53.5 / 57.9 \\
                        
                         &P3D-63\cite{qiu2017learning}  &53.4\\
         UCF101-DVS  &C3D~\cite{tran2015learning}  &47.2 \\
                 & Res-SNN18 + RM ~\cite{yao2023sparser}   & 63.5   \\
                       & \textbf{TIM(Ours)} &\textbf{63.8} \\
        \hline
                       &C3D~\cite{tran2015learning}  &41.7 \\
                   
                        &P3D-63~\cite{qiu2017learning}
                        &40.4 \\
          HMDB51-DVS    &Res-SNN18 + RM ~\cite{yao2023sparser}   & 44.7\\ 
             &Spikepoint~\cite{ren2023spikepoint} &55.6\\
                       &\textbf{TIM(Ours)}  &\textbf{58.6} \\
        \hline
        \multirow{2}{*}{SHD} &Spikformer &85.1\\
          
                        &\textbf{TIM(Ours)} &\textbf{86.3} \\
        \hline
    \end{tabular}
    }
    \caption{Comparison with other benchmark results on UCF101-DVS, HMDB51-DVS and SHD.}
    \label{tab:udfhmdb}
\end{table*}
\subsection{UCF101-DVS And HMDB51-DVS}
The \textbf{UCF101-DVS} and \textbf{HMDB51-DVS} datasets represent neuromorphic adaptations of the well-known UCF101 and HMDB51 datasets, respectively. These adaptations transform the original video-based datasets into a neuromorphic format, creating a rich collection of visual event streams. The UCF101-DVS encompasses an array of 101 action categories, while the HMDB51-DVS includes 51 distinct action categories. In these datasets, each action category is meticulously converted into dynamic visual event streams using advanced event cameras, offering a novel perspective on action recognition. Our TIM has showcased exceptional performance on these challenging datasets. On the HMDB51-DVS dataset, TIM achieved a top-1 accuracy of 58.6\%, and on the UCF101-DVS dataset, it reached a top-1 accuracy of 63.8\%, as detailed in Tab.~\ref{tab:udfhmdb}. 
\begin{table}[h]
    \centering
    \resizebox{0.4\textwidth}{!}{
    \begin{tabular}{llll}
        \hline
        \textbf{Dataset}  & \textbf{Model} &\textbf{Acc@1(\%)}  \\ 
        \hline
                        
         \multirow{2}{*}{CIFAR10-DVS} & Baseline  &78.5\\
            
                        &Baseline + \textbf{TIM}  &81.6 \\
        \hline
        \multirow{2}{*}{N-CALTECH101} &Baseline  &77.7\\
          
                        &Baseline + \textbf{TIM } &79.0 \\
        \hline
    \end{tabular}
    }
    \caption{Ablation study of the temporal interaction module.}
    \label{tab:ablation}
\end{table}
\subsection{SHD}
The Spiking Heidelberg Digits (\textbf{SHD}) dataset is an audio-based classification dataset containing 1000 spoken digits ranging from zero to nine in both English and German languages, with a total of 20 classes. Speech datasets typically exhibit more complex interdependence between time steps. TIM demonstrates a 1.2\% advantage over the baseline model on this dataset, shown in Tab~\ref{tab:udfhmdb}. The results indicates TIM's capability to extract intricate temporal features.

\section{Discussion}
\begin{figure}[ht]
  \centering
  \graphicspath{{images/}}
  \includegraphics[width=0.5\textwidth]{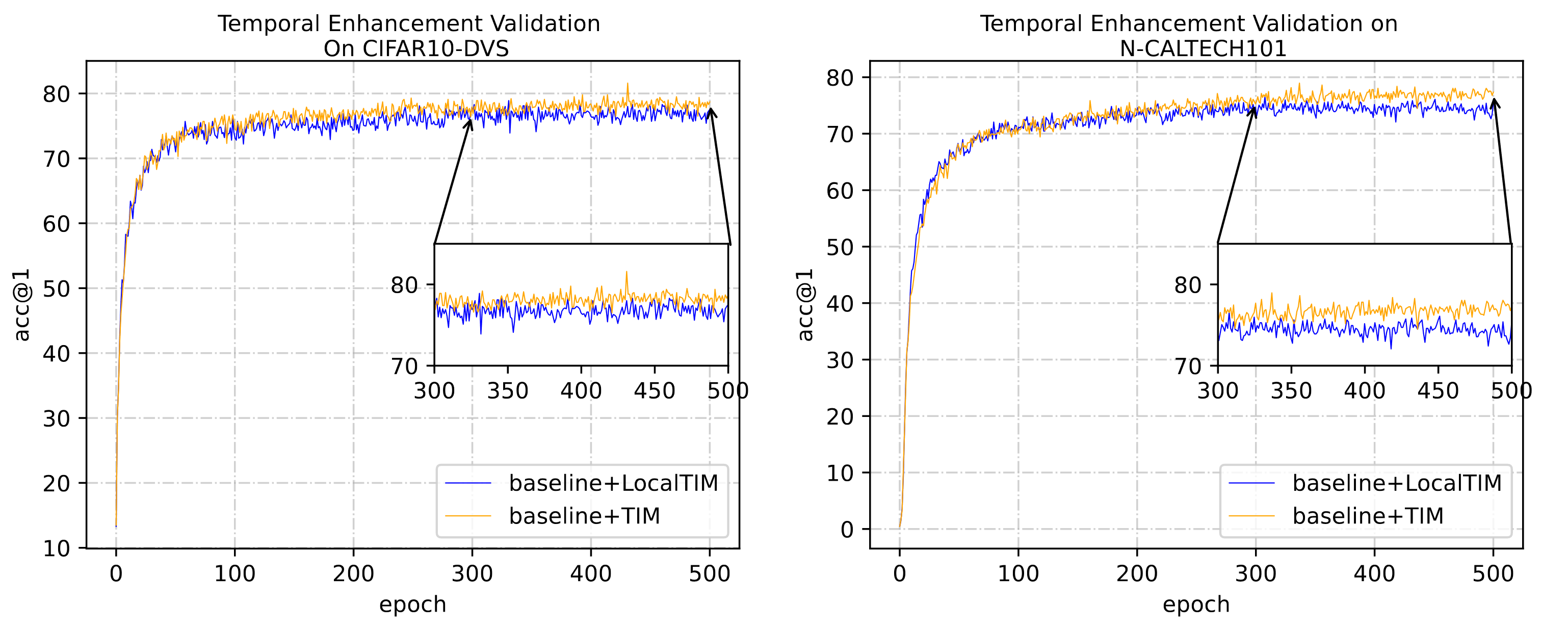}
  \caption{Temporal enhancement validation of TIM on CIFAR10-DVS and N-CALTECH101.}
  \label{fig:cr1}
\end{figure}
\subsection{Ablation Study}

To thoroughly validate the effectiveness of our algorithm, we embarked on comprehensive ablation studies utilizing the CIFAR10-DVS and N-CALTECH101 datasets. These datasets, known for their complexity and diverse visual event representations, provided an ideal testing ground for assessing the capabilities of our model. As detailed in Tab.~\ref{tab:ablation} and Fig.~\ref{fig:cr1}, our model demonstrated substantial performance enhancements over the established baseline models in both datasets.

\subsection{Temporal Enhancement Validation}
To rigorously validate that the performance enhancements observed in the Spiking Transformer are indeed due to its improved temporal processing capabilities, and not merely a consequence of an increased parameter count, we undertook a meticulous adjustment of the Temporal Interaction Module's structure. This refinement was crucial in isolating the impact of temporal capability enhancements from the effects of parameter augmentation. As detailed in Eq.~\eqref{eq:ablation1} and \eqref{eq:ablation2}, we specifically re-engineered the operational mechanism of the TIM Stream. This modification was aimed at changing the functional dynamics of the TIM Unit. Instead of allowing the TIM Unit to engage in interactions across multiple time steps, which could potentially confound our assessment of its temporal processing proficiency, we constrained its operation to the current time step only (which we call local TIM).

\begin{equation}\label{eq:ablation1}
    Q[t] =LIFNode(TIM(Q[t]]))
\end{equation}
\begin{equation}\label{eq:ablation2}
    Attention[t] = Q[t]K^T[t]V[t]
\end{equation}
As illustrated in Tab.~\ref{tab:ablation_localTIM}, a notable experiment was conducted applying our model to the CIFAR10-DVS and N-CALTECH101 datasets. The results were quite revealing: the model's accuracy decreased by 2.7\% and 2.4\% on these datasets, respectively. This phenomenon strongly indicates that the improvement in accuracy previously attributed to the Temporal Interaction Module is primarily a result of its enhanced ability to process temporal information, rather than the increased parameter count. The stability of the parameter count during the experiment reinforces this conclusion, highlighting the intrinsic value of TIM in specifically enhancing temporal data processing capabilities. The comprehensive training details are presented in Fig.~\ref{fig:cr1}.
\begin{table}[t]
    \centering
    \resizebox{0.43\textwidth}{!}{
    \begin{tabular}{llll}
        \hline
        \textbf{Dataset}  & \textbf{Model} &\textbf{Acc@1(\%)}  \\ 
        \hline
                        
         \multirow{2}{*}{CIFAR10-DVS} & Baseline+local TIM  &78.9\\
            
                        &Baseline + TIM  &81.6 \\
        \hline
        \multirow{2}{*}{N-CALTECH101} &Baseline+local TIM  &76.6\\
          
                        &Baseline + TIM  &79.0 \\
        \hline
    \end{tabular}
    }
    \caption{Temporal Enhancement Validation on CIFAR10-DVS and N-CALTECH101 datasets.}
    \label{tab:ablation_localTIM}
\end{table}
\begin{table}[t]
    \centering
    \resizebox{0.49\textwidth}{!}{
    \begin{tabular}{llllllll}
        \hline
        \textbf{alpha} & \textbf{0} &\textbf{0.2} &\textbf{0.4} &\textbf{0.5}  &\textbf{0.6} &\textbf{0.8} \\ 
        \hline
                        
         CIFAR10-DVS &78.5 &79.8 &79.7 &80.7 &\textbf{81.6} &81.2\\
        \hline
        N-CALTECH101 &77.7  &78.5 &\textbf{79.0} &78.5 &77.6 &78.7 \\
        \hline
    \end{tabular}
    }
    \caption{The top-1 accuracy of N-CALTECH101 and CIFAR10-DVS with different alpha values.}
    \label{tab:diff alpha}
\end{table}
We further concentrated on investigating the impact of the hyperparameter $\alpha$ on the performance of the Temporal Integration Module. This exploration was conducted through a series of experiments using the CIFAR10-DVS and N-CALTECH101 datasets. The top-1 accuracy of TIM on both datasets with different $\alpha$ values are illustrated in Tab.~\ref{tab:diff alpha}. The findings from these experiments were quite revealing. We observed that regardless of the specific value, setting $\alpha$ to any non-zero number consistently resulted in better performance compared to the baseline scenario where $\alpha$ was set to zero. This pattern indicates that the introduction of temporal interaction in TIM significantly enhances its performance. More specifically, for the CIFAR10-DVS dataset, the model's peak performance, with an accuracy of 81.6\%, was achieved when $\alpha$ was set to 0.6. Similarly, for the N-CALTECH101 dataset, the optimal performance occurred at an $\alpha$ value of 0.4, reaching a top accuracy of 79.0\%. These results highlight an important aspect of TIM's functionality. While the model exhibits robustness to a range of $\alpha$ values, fine-tuning this hyperparameter allows for more precise optimization relative to specific datasets. The ability to adjust $\alpha$ effectively tailors the model to different data distributions, enhancing TIM's versatility and efficacy in diverse neuromorphic data processing applications.

\subsection{Efficiency and Generalizability Validation}
\begin{figure}[h]
  \centering
  \graphicspath{{images/}}
  \includegraphics[width=0.49\textwidth]{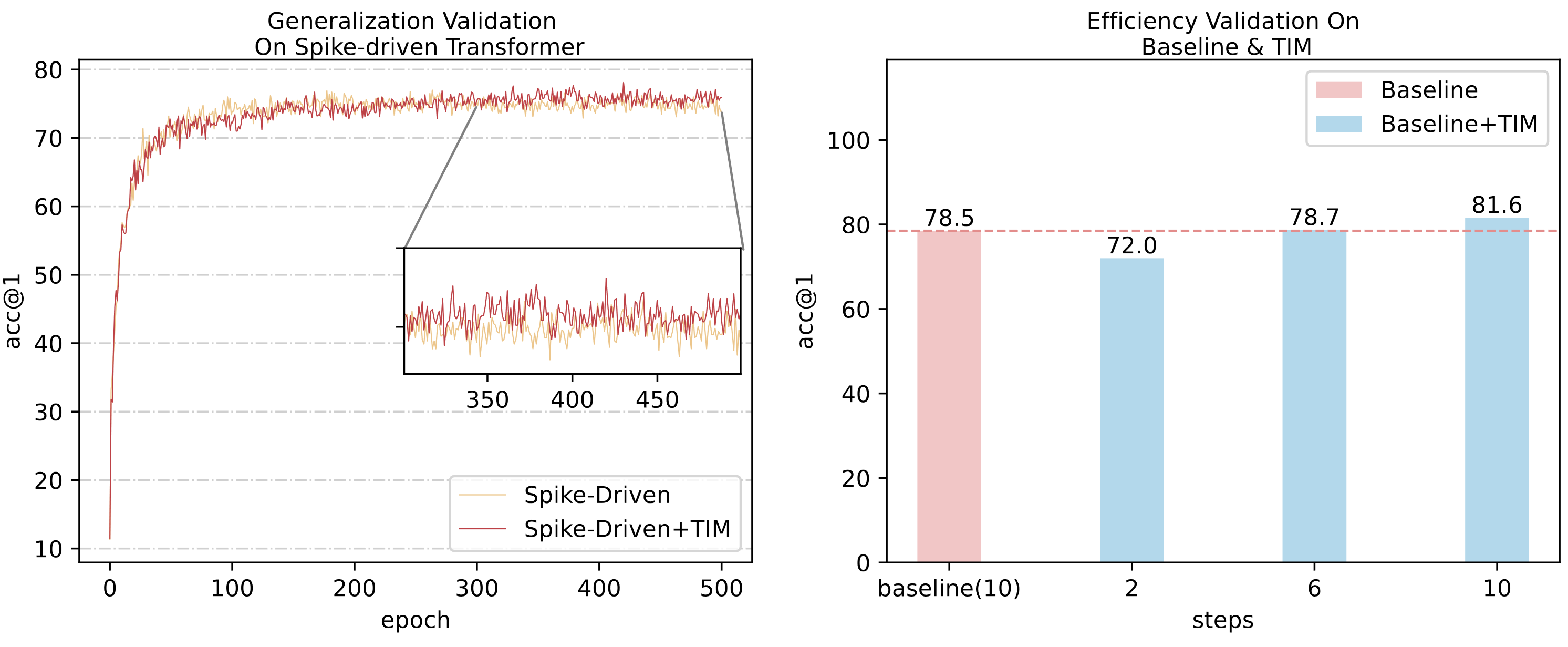}
  \caption{Efficiency and Generalizability Validation of TIM on CIFAR10-DVS.}
  \label{fig:cr2}
\end{figure}
TIM introduces additional operations into the model, resulting in extra computational overhead. To alleviate concerns about TIM's reduced efficiency, we reduced the training steps. We found that with 6 steps, TIM achieves the same performance as the baseline does with 10 steps, shown in Fig~\ref{fig:cr2}. This implies that TIM can save approximately 40\% of the time, suggesting that TIM retains the efficiency of SNN.\par

Furthermore, to validate the generalizability of TIM's capabilities to other Spiking Transformers, we conducted experiments on the Spike-driven Transformer as well. Before computing the Spike Driven Self Attention (SDSA, shown in Eq.~\ref{eq:SDSA})~\cite{yao2023spikedriven}, where $\odot$ refers to element-wise multiplication while $\textcircled{\scriptsize $\times$}$ denotes Hadamard product. We applied the same procedure to let Q pass through TIM, resulting in the curve shown in Fig~\ref{fig:cr2}.
\begin{equation}
\left\{
    \begin{aligned}
        Q[t],K[t],V[t] &= LIFNode(X[t]) \\
        A[t] &= LIFNode(K[t] \odot V[t])\\
        Attention[t] &= A[t] \textcircled{\scriptsize $\times$} Q[t]
    \end{aligned}
\right.
\label{eq:SDSA}
\end{equation}
SDSA achieved a top-1 accuracy of 77\% on CIFAR10-DVS with 10 steps, while SDSA+TIM achieved a top-1 accuracy of 78.5\%. We consider the 1.5\% difference to be statistically significant, indicating that TIM gains traction in SDSA, demonstrating its ability to generalize across a broader range of Spiking Transformer architectures.
\section{Conclusion}
In our research, we conduct a detailed examination of the current Spiking Transformer models, focusing particularly on their suboptimal use of temporal information. To address this issue, we introduce the Temporal Interaction Module (TIM), a groundbreaking, plug-and-play component designed to enhance the Spiking Transformer's ability to process temporal data more effectively. This module, skillfully constructed using one-dimensional convolutions, is notable for its minimal parameter count. Our methodology ensures simplicity in implementation while adhering to the Spiking Transformer's principles of efficiency and compactness. Rigorous experimental validations underscore the efficacy of TIM, revealing its superior performance, especially prominent in neuromorphic dataset applications. TIM not only excels in these contexts but also establishes new state-of-the-art benchmarks. Furthermore, Our discussion and ablation study demonstrate that TIM effectively extracts temporal information on complex datasets, leading to improvements in model performance. Meanwhile, TIM retains the efficiency characteristic of SNNs and the ability to generalize across various Spiking Transformer architectures.

\section*{Acknowledgments}
This research is financially supported by a funding from Institute of Automation, Chinese Academy of Sciences (Grant No. E411230101).

\bibliographystyle{named}
\bibliography{ijcai23}

\end{document}